\documentclass{article}

\usepackage[preprint]{colm2026_conference}

\usepackage[utf8]{inputenc}
\usepackage[T1]{fontenc}
\usepackage{amsmath,amssymb,amsfonts}
\usepackage{graphicx}
\usepackage{booktabs}
\usepackage{algorithm}
\usepackage{algorithmic}
\usepackage{hyperref}
\usepackage{xcolor}
\usepackage{multirow}
\usepackage{subcaption}
\usepackage{tikz}
\usetikzlibrary{arrows.meta,positioning,shapes.geometric,fit,calc}
\usepackage[margin=1in]{geometry}
\usepackage{natbib}
\usepackage{enumitem}
\usepackage{tcolorbox}
\usepackage{tabularx}
\usepackage{array}
\usepackage{makecell}
\usepackage{graphicx}
\usepackage{wrapfig}
\usepackage{caption}

\usepackage{pifont}
\newcommand{\cmark}{\checkmark}
\newcommand{\xmark}{\texttimes}
\newcommand{\pmark}{\(\triangle\)}

\usepackage{microtype}
\usepackage{url}

\usepackage{alltt}
\usepackage[T1]{fontenc}

\usepackage{lineno}

\hypersetup{
  colorlinks=true,
  linkcolor=blue!60!black,
  citecolor=blue!60!black,
  urlcolor=blue!60!black,
}

\newcommand{\method}{\textsc{ScioMind}}

\newcommand{\eg}{\textit{e.g.}}

\definecolor{darkblue}{rgb}{0, 0, 0.5}
\hypersetup{colorlinks=true, citecolor=darkblue, linkcolor=darkblue, urlcolor=darkblue}

\title{\method{}: Cognitively Grounded Multi-Agent Social Simulation with Anchoring-Based Belief Dynamics and Dynamic Profiles}

\author{
  Yitian Yang$^{1}$  \\
  The University of Sydney \\
  \texttt{yitian.yang@sydney.edu.au} \\
  \And
  Yiqun Duan$^{1,*}$  \\
  Meta\\
  \texttt{duanyiquncc@gmail.com} \\
  \And
  Linghan Huang$^{2}$ \\
  The University of Sydney \\
  \texttt{linghan.huang@sydney.edu.au} \\
  \And
  Yiqi Zhu$^{2}$ \\
  The University of Sydney \\
  \texttt{yzhu0325@uni.sydney.edu.au} \\
  \And
  Francesco Bailo$^{2}$ \\
  The University of Sydney \\
  \texttt{francesco.bailo@sydney.edu.au} \\
  \And
  Chunmeizi Su$^{1,\dagger}$ \\
  The University of Sydney \\
  \texttt{chunmeizi.su@sydney.edu.au} \\
  \And
  Huaming Chen$^{1,\dagger}$ \\
  The University of Sydney \\
  \texttt{huaming.chen@sydney.edu.au} \\
  \\
  $^{*}$Equal contribution. \quad
  $^{\dagger}$Corresponding authors.
}

\begin{document}


\maketitle

\begin{abstract}
Large language model (LLM)-based multi-agent simulation offers a powerful testbed for studying social opinion dynamics. Yet current approaches often adopt two contrasting methods: either relying on fixed update rules with limited cognitive grounding or delegating belief change largely to unconstrained LLM interaction. We introduce \method{}, a cognitively grounded simulation framework that bridges these paradigms by combining structured opinion dynamics with LLM-based agent reasoning. \method{} integrates three key components: 1) a memory-anchored belief update rule that modulates susceptibility to influence via personality-conditioned anchoring strength; 2) a hierarchical memory architecture that supports persistent, experience-driven belief formation; and 3) dynamic agent profiles derived from a corpus-grounded retrieval pipeline, enabling heterogeneous personalities, rationales, and evolving internal states. We evaluate \method{} on multiple case studies in a real-world policy debate scenario. Across metrics including polarisation, diversity, extremization, and trajectory stability, the proposed components consistently yield improvements in behavioural realism. In particular, dynamic profiles increase opinion diversity, memory and reflection reduce unstable oscillation, and anchoring induces persistent belief trajectories that better align with patterns reported in political psychology. These results suggest that our cognitively grounded design provides a novel solution to LLM-based social simulation that improves both stable and behavioural realism.
\end{abstract}

\section{Introduction}
\label{sec:intro}

In computational social science, understanding how beliefs form, persist, and evolve in social networks remains a central challenge.
Classical agent-based models (ABMs) such as the DeGroot consensus model~\citep{degroot1974reaching}, bounded confidence models~\citep{deffuant2000mixing,hegselmann2002opinion}, and Axelrod's culture dissemination model~\citep{axelrod1997dissemination}, have provided foundational insights into opinion dynamics.
However, these models generally rely on simplified update rules, often based on weighted averaging of scalar opinions, which limits their ability to capture the nuanced reasoning process, persuasive interactions, and belief updating process that characterize real human belief formation.

The emergence of large language models (LLMs) has opened new possibilities for simulating opinion evolution in social systems.
Generative Agents~\citep{10.1145/3586183.3606763} demonstrated that memory and reflection can support believable social behaviour, while subsequent systems,including S$^3$~\citep{gao2023s3}, AgentSociety~\citep{piao2025agentsociety}, and OASIS~\citep{yang2024oasis}, have extended LLM-based social simulation to larger populations.
Compared with traditional ABMs, these approaches improve the expressiveness and interpretability of how opinions and beliefs evolve through interaction.
However, they typically treat belief updating as an emergent outcome of open-ended interaction, which introduces several limitations.
First, these methods lack explicit and structured representations of belief influence dynamics, making it difficult to identify what drives an agent's opinion update.
Second, they do not incorporate cognitively grounded mechanisms of resistance to change.
Third, agent diversity is often limited to static demographic or role-based descriptions, rather than dynamically evolving internal traits, rationales and memories.
Finally, on controversial topics, these systems often exhibit over-smoothing, in which agents normally converge too quickly toward neutral or consensual positions, failing to preserve the persistent disagreement and bounded polarisation commonly observed in real human debates.

Our work is motivated by the \emph{anchoring effect}, a well-established concept in cognitive psychology and opinion change research, whereby judgments are disproportionately shaped by initial reference points and adjusted only insufficiently thereafter~\citep{tversky1974judgment}.
Recent work suggests that this mechanism is particularly salient in political preference formation, where the status quo can act as a cognitive anchor that resists change even under persuasive pressure~\citep{arceneaux2022anchoring}.
Importantly, these effects are stronger when anchors are \emph{personally relevant}, such as existing policy positions, suggesting that belief persistence arises not from arbitrary inertia but from accumulated experience.
This observation directly motivates our core design of a \emph{memory-anchored} belief update mechanism, where each agent's resistance to opinion change is shaped by experience-based anchors maintained in long-term memory.

\paragraph{Contributions.}
We introduce \method{}, a cognitively grounded framework for LLM-based social simulation with four main contributions:

\begin{enumerate}[leftmargin=*,itemsep=2pt]
    \item \textbf{Anchoring-based belief dynamics} (\S\ref{sec:anchoring}): We model belief updating as a memory-anchored influence process, where agents revise opinions related to topic-specific anchors derived from memory. Anchoring strength is mapped from agents' personality traits, introducing cognitively grounded heterogeneity in persuadability.

    \item \textbf{Cognitive memory architecture} (\S\ref{sec:memory}): We design a four-layer memory system comprising episodic, semantic, reflection, and working memory, which supports the formation and retrieval of belief anchors and explicitly links memory mechanisms to belief dynamics.

    \item \textbf{Dynamic agent profiles} (\S\ref{sec:profiles}): We replace static demographic profiles with narrative identities retrieved from a social-media-grounded corpus, combined with group-conditioned OCEAN priors and rationale clusters to produce diverse agents with evloving and internally grounded profiles.


    \item \textbf{Human-in-the-loop simulation design and evaluation}: We incorporate domain expert feedback to calibrate anchoring strength and bounded confidence, using observed patterns such as polarisation, clustering, and convergence to iteratively improve the simulation's behavioural realism and interpretability.

\end{enumerate}

We evaluate these contributions through case studies on real-world policy debate scenarios, assessing both component effects and the extent to which the agents' behaviours align with human patterns.

\section{Related Work}
\label{sec:related}

\paragraph{Classical opinion dynamics.}
Classical opinion-dynamics research studies how individual beliefs evolve through repeated social interaction on a network, typically using explicit update rules over low-dimensional opinion states. One foundational influence dynamic model is the DeGroot model~\citep{degroot1974reaching}, which formalizes opinion formation as iterative weighted averaging. Building on this, the Friedkin--Johnsen model~\citep{friedkin1990social} introduces \emph{stubbornness}, whereby agents retain partial attachment to their initial opinions, providing a mathematical precursor to our anchoring formulation.
Bounded confidence models~\citep{deffuant2000mixing,hegselmann2002opinion} further show that how polarisation can emerge through selective interaction.
Despite mathematically tractable, these models operate over scalar opinions states and offer limited ability to capture the argumentative richness involved in human belief formation.


\paragraph{LLM-based social simulation.}
The emergence of LLMs has made social simulation a promising direction for modelling complex human behavior. Compared with traditional agent-based modelling, LLM-based simulations enable more behaviourally realistic agents, providing interpretable rationales for agents' actions beyond numerical update rules, and supporting reproducible experimental settings. \citet{10.1145/3586183.3606763} introduces generative agents with observation--reflection--planning loops. S$^3$~\citep{gao2023s3} models social networks with emotion tracking. OASIS~\citep{yang2024oasis} scales simulations to one million agents. AgentSociety~\citep{piao2025agentsociety} examines polarisation with LLM agents, while MOSAIC~\citep{liu2025mosaic} focuses on simulating belief dynamics in misinformation communication. \citet{yazici2026opinion} shows that LLM agents operating under DeGroot protocols exhibit convergence patterns consistent with graph-theoretic predictions, albeit with topic-dependent biases. \citet{nasim2025simulating} further investigates the dynamics of social influence. However, existing LLM-based social simulation frameworks still lack an explicit, cognitively grounded account of how beliefs persist and update through social interaction, particularly in ways shaped by opinion dynamics, diverse personality, and prior experience. Table~\ref{tab:comparison} positions \method{} relative to prior work across six dimensions. Our framework provides a novel solution to enable more behaviourally grounded and structurally explicit simulations of social opinion dynamics.

\begin{table}[t]
\centering
\caption{Comparison of LLM-based social simulation systems. 
\textbf{Belief}: structured belief states; 
\textbf{Mem.}: memory mechanism; 
\textbf{Dyn. Prof.}: dynamic agent profiles; 
\textbf{Refl.}: reflection mechanism; 
\textbf{Topo.}: configurable topologies; 
\textbf{KOL-Net}: KOL-aware relationship or influence-network construction; 
\textbf{Scen. Samp.}: scenario-aligned sampling; 
\textbf{Bal. Cohort}: balanced stance cohort for comparative analysis.}
\label{tab:comparison}
\small
\resizebox{\textwidth}{!}{
\begin{tabular}{lcccccccc}
\toprule
\textbf{System} & \textbf{Belief} & \textbf{Mem.} & \textbf{Dyn. Prof.} & \textbf{Refl.} & \textbf{Topo.} & \textbf{KOL-Net} & \textbf{Scen. Samp.} & \textbf{Bal. Cohort} \\
\midrule
Generative Agents~\citeyearpar{10.1145/3586183.3606763} 
& \xmark & \xmark & \xmark & \cmark & \xmark & \xmark & \xmark & \xmark \\

S$^3$~\citeyearpar{gao2023s3} 
& \xmark & \xmark & \xmark & \xmark & \xmark & \xmark & \xmark & \xmark \\

OASIS~\citeyearpar{yang2024oasis} 
& \xmark & \xmark & \xmark & \xmark & \xmark & \xmark & \xmark & \xmark \\

AgentSociety~\citeyearpar{piao2025agentsociety} 
& \xmark & \cmark & \xmark & \xmark & \xmark & \xmark & \xmark & \xmark \\

MOSAIC~\citeyearpar{liu2025mosaicmodelingsocialai} 
& \xmark & \cmark & \cmark & \xmark & \cmark & \pmark & \xmark & \xmark \\

TinyTroupe~\citeyearpar{salem2025tinytroupe} 
& \xmark & \xmark & \xmark & \xmark & \xmark & \xmark & \xmark & \xmark \\

SPARK~\citeyearpar{zhang2025spark} 
& \cmark & \cmark & \xmark & \xmark & \xmark & \xmark & \xmark & \xmark \\

POSIM~\citeyearpar{zhang2026posim} 
& \cmark & \pmark & \xmark & \xmark & \pmark & \pmark & \xmark & \xmark \\

\midrule
\method{} (Ours) 
& \cmark & \cmark & \cmark & \cmark & \cmark & \cmark & \cmark & \cmark \\
\bottomrule
\end{tabular}
}
\end{table}

\section{Methodology}
\label{sec:method}

\begin{figure*}[t]
    \centering
    \begin{subfigure}[t]{\textwidth}
        \centering
        \includegraphics[width=\textwidth]{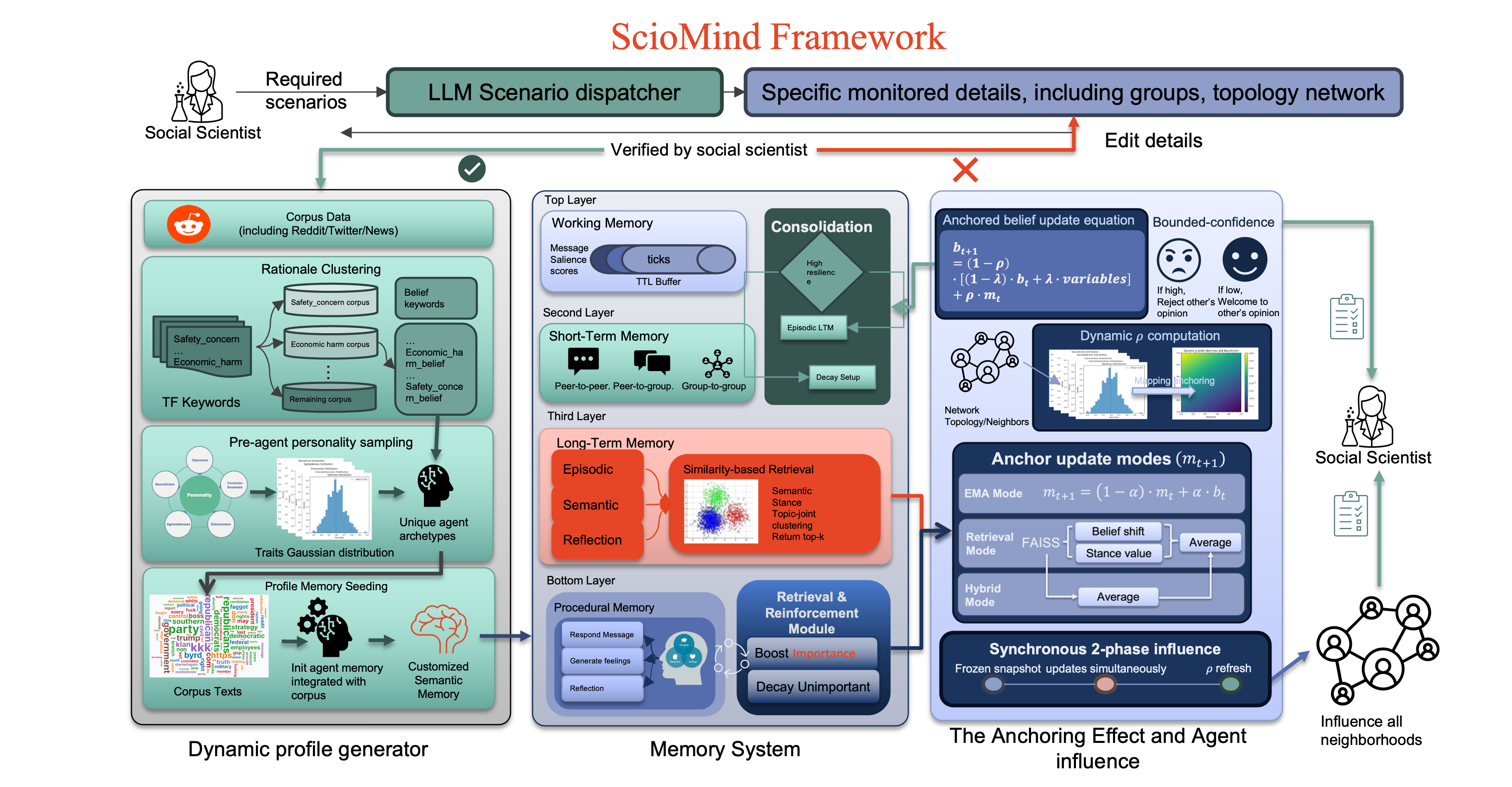}
        \label{fig:overview}
    \end{subfigure}
    \caption{
    Architecture overview. 
    }
    \label{fig:overview_diagram}
\end{figure*}

\method{} simulates belief dynamics across heterogeneous social groups through a layered architecture (Figure~\ref{fig:overview_diagram}). We begin by describing the framework's three core contributions, followed by the simulation loop with human-in-the-loop for calibration and evaluation.








\subsection{Anchoring-Based Belief Dynamics}
\label{sec:anchoring}

Inspired by the anchoring effect in political judgment~\citep{arceneaux2022anchoring,tversky1974judgment}, we model resistance to belief change as attachment to a \emph{memory anchor}, which is a reference point derived from the agent's accumulated experience.

\paragraph{Anchored belief update.}
For agent $i$ with belief $b_i^{k,t} \in [-1,1]$ on topic $k$ at tick $t$, neighbors $\mathcal{N}_i$ with trust weights $w_{ij}$, and memory anchor $m_i^{k,t}$, the belief update rule is:
\begin{equation}
    b_i^{k,t+1} = \underbrace{(1 - \rho_i)}_{\text{social openness}} \Big[\underbrace{(1 - \lambda_i) \, b_i^{k,t}}_{\text{self-weight}} + \underbrace{\lambda_i \, S_i^{k,t}}_{\text{social influence}}\Big] + \underbrace{\rho_i \, m_i^{k,t}}_{\text{anchor pull}},
    \label{eq:anchored_degroot}
\end{equation}
where $\lambda_i \in [0,1]$ denotes \emph{susceptibility} and $\rho_i \in [0,1]$ is the \emph{anchoring strength}.

The social influence term incorporates trust-weighted influence with belief homophily~\citep{aybas2026social} under bounded confidence:
\begin{equation}
    S_i^{k,t} = \sum_{j \in \mathcal{N}_i} \hat{w}_{ij} \cdot 
    \varphi_{ij}^t \cdot b_j^{k,t}, \quad
    \hat{w}_{ij} = \frac{w_{ij}}{\sum_{\ell} w_{i\ell}}, \quad
    \varphi_{ij}^t = \max\!\big(0,\; 1 - |b_i^{k,t} - b_j^{k,t}|\big).
    \label{eq:social_term}
\end{equation}

Here, $\varphi_{ij}^t$ represents belief homophily. When $\rho_i = 0$, Eq.~\ref{eq:social_term} recovers a standard bounded-confidence DeGroot model.
While $\rho_i > 0$ with $m_i = b_i^{k,0}$ (fixed initial anchor), it recovers a form of the Friedkin--Johnsen stubbornness model~\citep{friedkin1990social}.
Our key extension is that $m_i^{k,t}$ is \emph{dynamic} and evolves with the agent's experience. This allows the anchor to remain personally relevant over time, featured by~\citet{arceneaux2022anchoring}, reflecting that anchors grounded in personal experience exert stronger influence than arbitrary reference points.

\paragraph{Personality and sentiment based anchoring strength}
Rather than assigning $\rho_i$ uniformly, we derive it from the agent's Big Five personality profile via a sigmoid mapping:
\begin{equation}
    \rho_i = \rho_{\min} + (\rho_{\max} - \rho_{\min}) \cdot \sigma\!\big(\gamma_0 + \gamma_O \cdot O_i + \gamma_C \cdot C_i + \gamma_E \cdot E_i + \gamma_A \cdot A_i + \gamma_N \cdot N_i\big),
    \label{eq:ocean_to_rho}
\end{equation}
where $\sigma(\cdot)$ is the logistic function, $\rho_{\min}=0.05$, $\rho_{\max}=0.5$, and the coefficients $\gamma$ encode psychological priors: low Openness ($\gamma_O < 0$) and high Conscientiousness ($\gamma_C > 0$) increase anchoring strength, reflecting that closed-minded and methodical individuals resist belief change more strongly.




\subsection{Cognitively Grounded Memory Architecture}
\label{sec:memory}

Each agent maintains a four-namespace memory system inspired by cognitive science BDI models of human memory~\citep{hu2025memory}:

\paragraph{Episodic memory (short-term memory).}
A fixed-capacity sliding window of recent interaction messages.
This provides immediate conversational context, analogous to working memory. Each agent's episodic memory is maintained as a FIFO deque.

\paragraph{Semantic memory (long-term, corpus-seeded).}
An unbounded store of \texttt{ProfileMemoryEntry} records, each containing content, topic, stance, rationale tag, source platform, and importance score.
Entries are indexed in a FAISS vector store~\citep{johnson2021faiss} using text embeddings, enabling efficient semantic retrieval.
At initialisation, semantic memory is \emph{seeded} from a cleaned customised corpus dataset through RAG (\S\ref{sec:profiles}), giving each agent a unique experiential history.

At each cognitive loop iteration, the top-$k$ ($k=4$) most semantically similar memories are retrieved and injected into the LLM reasoning prompt.
The agent's accumulated experience serves as summarising knowledge base that grounds reasoning in personal history~\citep{lewis2020retrieval}.

\paragraph{Procedure memory. }

Formally, the procedural memory is defined as a set of routines $\mathcal{R} = \{r_1, r_2, \dots, r_K\}$. Each routine $r \in \mathcal{R}$ is characterized by a tuple:
\begin{equation}
    r = \langle C_r, \rho_r, \mathcal{M}_r \rangle
\end{equation}
where $C_r$ denotes the set of preconditions (e.g., \textit{has\_unread\_messages} = True) that must be satisfied for the routine to be executable; $\rho_r \in [0, 1]$ represents the base priority of the routine; and $\mathcal{M}_r: \mathcal{E} \rightarrow \mathbb{R}^+$ is an emotion modulation function that maps a discrete Ekman emotion $e \in \mathcal{E}$, such as joy, fear, angry to a scalar multiplier.

\paragraph{Reflection memory.}
After each tick, every agent generates a self-reflection summarizing received messages, computed belief shifts, and confidence level:
\begin{equation}
    \text{confidence}_i^t = 1.0 - \sum_{k} |\Delta b_i^{k,t}|, \quad \Delta b_i^{k,t} = b_i^{k,t} - b_i^{k,t-1}.
    \label{eq:confidence}
\end{equation}
Reflections are stored in the \texttt{reflection} namespace and serve dual purposes: (i)~they are retrieved during subsequent reasoning to provide self-awareness, and (ii)~they feed directly into the anchoring module's update, creating a closed loop where accumulated reflection strengthens belief anchors.

Combined with the memory part, we implement three anchor update strategies: \textbf{EMA}, which exponentially smooths prior anchors with current beliefs; \textbf{Retrieval}, which computes anchors from topic-relevant long-term memories; and \textbf{Hybrid}, which linearly combines the two to balance recency and memory grounding.

\subsection{Social Relationship Simulation Engine}
\label{sec:social-relationship-engine}

The Social Relationship Simulation Engine in this project is designed to integrate AI-generated individual profiles, profile-based social relationship mining, circle-aware network construction, and scenario-aligned candidate sampling into a unified simulation pipeline. The system first generates or reads a pool of agent candidate profiles. It then extracts relational features from demographic attributes, interests, stances, OCEAN personality traits, group identities, and semantic tags to construct latent social circles and directed trust relations among agents. Finally, it performs scenario-specific sampling and network initialization for each simulation run.

\subsubsection{Dynamic Agent Profile Generation}

Let the candidate profile pool be
\[
\mathcal{P}=\{p_i\}_{i=1}^{N},
\]
where each profile $p_i$ contains demographic attributes $d_i$, a personality vector $o_i$, an interest set $I_i$, initial beliefs $b_i$, a textual description $m_i$, and metadata $z_i$. In the AI-assisted generation mode, the system uses the group prompt, case-study context, demographic constraints, and summaries of existing profiles as conditional inputs, and calls an LLM to generate structured JSON profiles in batches. The system then applies deterministic normalization to reduce repetition and improve controllability in LLM-generated outputs.

Demographic constraints are approximated using the largest remainder method. If the target proportion of category $c$ in a given field is $\pi_c$ and the target sample size is $N$, then the target count for category $c$ is approximated as
\[
n_c=\lfloor N\pi_c \rfloor + \mathbb{I}\{c \in \mathcal{R}\},
\]
where $\mathcal{R}$ denotes the set of categories that receive the remaining slots after sorting by fractional remainders. Interest assignment is further constrained by limiting the overlap between any two profiles:
\[
|I_i \cap I_j| \leq \kappa,
\]
where $\kappa$ is the maximum interest-overlap threshold. The system also enforces unique names and biographies, and rewrites or supplements profile descriptions when necessary to avoid template-like profiles.

If OCEAN traits are not explicitly provided, the system samples personality values from group-specific archetype priors:
\[
o_{ik} \sim \operatorname{clip}\left(\mathcal{N}(\mu_{g_i k}, \sigma_{g_i k}^{2}), 0.02, 0.98\right),
\]
where $g_i$ denotes the group to which agent $i$ belongs, and $k$ indexes the OCEAN dimensions. This design preserves within-group personality heterogeneity while retaining group-level behavioral tendencies.

Generation quality is constrained using a profile diversity score:
\[
D = 0.28U_{\text{name}} + 0.32U_{\text{bio}}
+0.20\left(1-\min\left(1,\frac{\overline{O}_{I}}{3}\right)\right)
+0.20(1-\overline{J}_{\text{bio}}),
\]
where $U_{\text{name}}$ and $U_{\text{bio}}$ denote the proportions of unique names and unique biographies, respectively; $\overline{O}_{I}$ is the average number of overlapping interests; and $\overline{J}_{\text{bio}}$ is the average Jaccard similarity between biography texts.

\paragraph{Per-agent OCEAN sampling.}
Rather than assigning all agents in a group the same OCEAN vector, we sample each agent's traits from group-specific Gaussian priors:
\begin{equation}
    O_i \sim \mathcal{N}(\mu_O^g, \sigma_O^g), \quad C_i \sim \mathcal{N}(\mu_C^g, \sigma_C^g), \quad 
    E_i \sim \mathcal{N}(\mu_E^g, \sigma_E^g), \quad
    A_i \sim \mathcal{N}(\mu_A^g, \sigma_A^g), \quad N_i \sim \mathcal{N}(\mu_N^g, \sigma_N^g)
    \label{eq:ocean_sampling}
\end{equation}
where $(\mu^g, \sigma^g)$ are calibrated per group (\eg, GovernmentGroup: $\mu_C = 0.8, \sigma_C = 0.08$; CitizenGroup: $\mu_C = 0.5, \sigma_C = 0.12$).
Values are clipped to $[0, 1]$.
This creates within-group personality diversity that propagates to heterogeneous anchoring strengths via Eq.~\ref{eq:ocean_to_rho}.

\subsubsection{Circle Construction and Relationship Mining}

The system defines a circle as a latent social grouping formed by shared social attributes, issue stances, interests, roles, or demographic characteristics, rather than as a closed cycle in the graph-theoretic sense. For each agent, the system first computes the average belief value:
\[
\bar{b}_i = \frac{1}{|\mathcal{T}_i|}\sum_{t\in \mathcal{T}_i} b_{it},
\]
and uses it to assign an initial stance:
\[
s_i =
\begin{cases}
\text{supportive}, & \bar{b}_i > 0.25,\\
\text{opposing}, & \bar{b}_i < -0.25,\\
\text{neutral}, & \text{otherwise}.
\end{cases}
\]

The system then constructs feature tags:
\[
T_i = \{\text{group}, \text{role}, \text{case}, \text{stance}, \text{topic},
\text{demographics}, \text{interests}, \text{rationale}\}.
\]
Based on these tags, the circle membership of agent $i$ is defined as
\[
C_i =
\{\texttt{community}:g_i,\ \texttt{issue}:t_i:s_i,\ \texttt{role\_circle}:r_i\}
\cup \{\texttt{circle}:x \mid x \in T_i\}.
\]
If an agent is identified as a key opinion leader (KOL), the system additionally assigns local-hub or cross-network-hub tags.

Inter-agent relationships are determined jointly by feature similarity and circle overlap:
\[
J_T(i,j)=\frac{|T_i\cap T_j|}{|T_i\cup T_j|}, \qquad
J_C(i,j)=\frac{|C_i\cap C_j|}{|C_i\cup C_j|}.
\]
For each directed candidate edge $(i,j)$, the system computes a relationship score:
\[
S_{ij}=A_c(g_i,g_j)
+\lambda_C J_C(i,j)
+\lambda_T J_T(i,j)
+0.16\mathbb{I}[\ell_i=\ell_j]
+\rho_{ij}
+\kappa_{j}
-\delta_{ij}.
\]
Here, $A_c(g_i,g_j)$ denotes the case-study-specific group affinity; $\ell_i$ is the latent block assignment; $\rho_{ij}$ is a reciprocity bonus; $\kappa_j$ is an additional weighting term for target nodes identified as KOLs or influencers; and $\delta_{ij}$ is a degree-penalty term. The system first establishes local KOL links, cross-group KOL meshes, and bridge edges, and then greedily fills each agent's target out-degree according to $S_{ij}$. Relationship types include \texttt{friend}, \texttt{colleague}, \texttt{follower}, \texttt{influencer}, and \texttt{teacher}. The corresponding trust weight is defined as
\[
w_{ij}=\operatorname{clip}\left(w_0(\tau_{ij})+0.15J_C(i,j)+0.10J_T(i,j),0.10,0.95\right).
\]

\subsubsection{Scenario-Aligned Sampling}

Before each simulation run, the system can perform scenario-aligned sampling over the candidate pool. Given a scenario title $q$ and a set of key topics $\mathcal{K}$, the relevance score of each profile is defined as
\[
R_i=\min\left(1,\ 0.15+
\frac{|V(q,\mathcal{K})\cap V(p_i)|}{|V(q,\mathcal{K})|}
+\eta_i\right),
\]
where $V(\cdot)$ denotes a set of textual tokens, and $\eta_i$ is a role-based bonus for brokers, experts, organizers, institutional actors, or KOLs.

The system first infers each agent's initial stance using, in order, explicit stance labels, mean belief values, and textual cues. If stance information is missing, it is imputed according to quota targets or a default balanced rule. The system then allocates sample sizes across groups:
\[
n_g \approx n \cdot \frac{|\mathcal{P}_g|}{|\mathcal{P}|},
\]
and applies the largest remainder method to ensure that the total sample size is preserved.

The current implementation supports multiple sampling strategies. \texttt{stratified} sampling stratifies agents by initial stance or a specified field; \texttt{quota} sampling enforces proximity to target stance proportions; \texttt{cluster} sampling selects highly relevant clusters based on circles, feature tags, or interests; \texttt{multistage} sampling first clusters agents and then applies stratification; \texttt{purposive} sampling jointly maximizes scenario relevance and novelty in roles and circles; \texttt{snowball} sampling expands from KOLs or specified seeds along the relationship network; and \texttt{theoretical} sampling iteratively covers underrepresented stances, groups, roles, and circles.

To provide a comparable reference population, the system also constructs a balanced \(1{:}1{:}1\) cohort. If supportive, neutral, and opposing agents are all present, let
\[
k=\min_{s\in\{\text{supportive},\text{neutral},\text{opposing}\}}|\mathcal{P}_s|.
\]
The balanced cohort is then formed by selecting the top-$k$ agents with the highest scenario relevance from each stance category. The final system logs the realized sample size, group distribution, initial stance distribution, average scenario relevance, and balanced cohort membership for subsequent causal analysis and historical replay.

\begin{figure*}[t]
    \centering
    \includegraphics[width=0.8\linewidth]{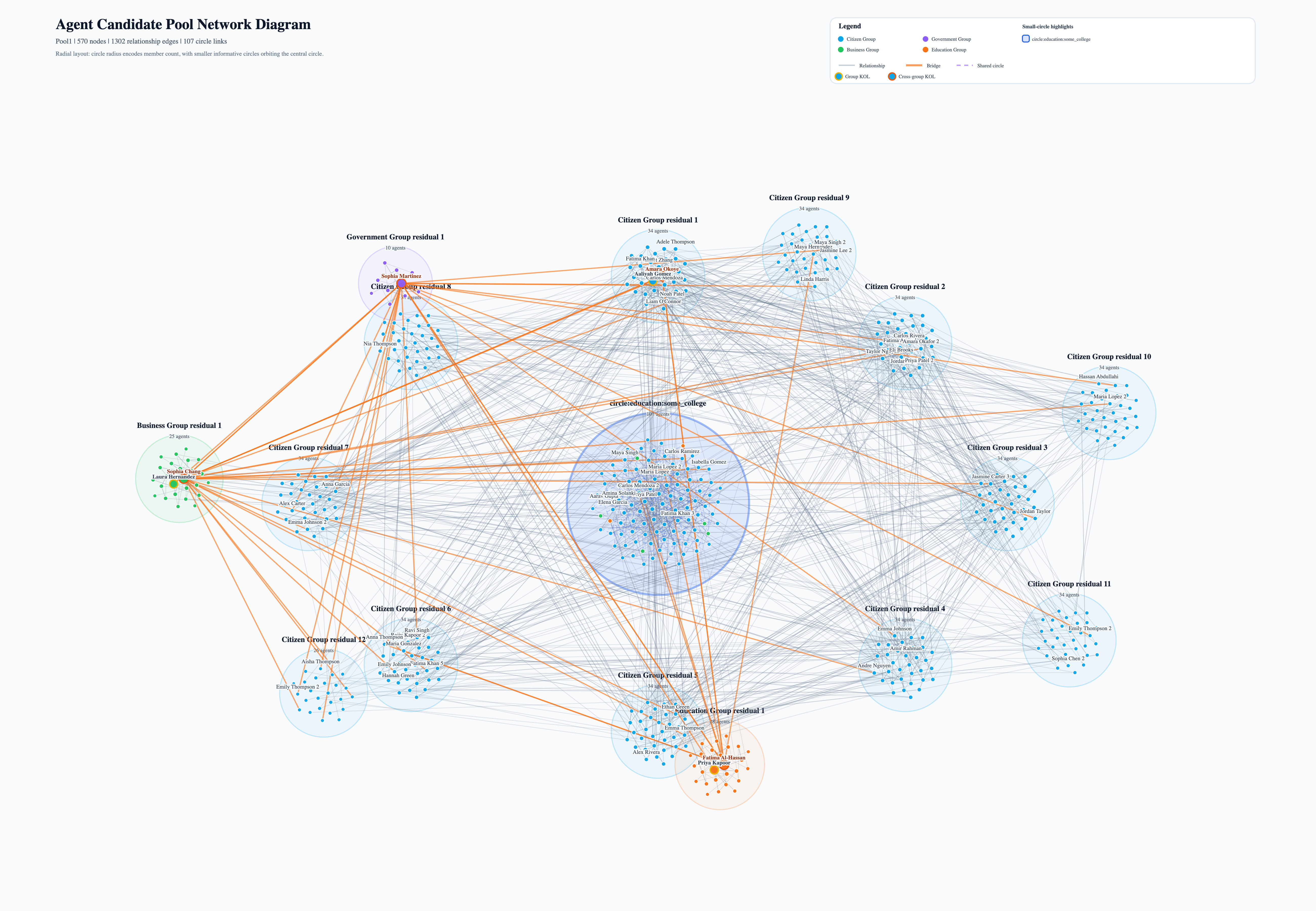}
    \caption{Agent's Social Network Visualisation}
    \label{fig:agent_social_network}
\end{figure*}

\subsection{Simulation Algorithm}
The detailed simulation flow is provided in Appendix~\ref{app:simulation_algorithm} Algorithm ~\ref {alg:simulation_flow}. Users specify the required inputs, such as simulation ticks number $T$, and verify that the LLM scenario dispatcher correctly maps the corresponding simulation configurations. We further provide an interactive simulation system for evaluation\footnote{\url{https://sciominds.vercel.app}}.

\subsection{Human-in-the-Loop}
Anchoring strength ($\rho$) and bounded confidence are central parameters governing interpersonal influence in the simulation. Bounded confidence determines whether and how agents incorporate neighbouring opinions: agents are more receptive to beliefs within a limited similarity range, while large belief discrepancies are discounted or rejected, giving rise to echo chamber effects and persistent opinion fragmentation.

We incorporate human-in-the-loop evaluation to calibrate these parameters against real-world social dynamics. Social science experts evaluate whether emergent patterns—such as polarisation levels, opinion clustering, and convergence behaviour—are consistent with empirical observations. This enables iterative refinement of the model to ensure behavioural realism and interpretability.

\section{Experiments}
\label{sec:experiments}

\subsection{Experimental Setup}
\label{sec:setup}

\subsubsection{Implementation Details}

We implement our ~\method{} simulation framework under python environment. LLM-based reasoning and scenario parsing are performed via the OpenAI API. Each agent is instantiated as one of four societal archetypes—Citizen, Government, Business, or Education—and parameterised by Big Five personality traits (OCEAN)~\citep{gerber2011big}, detailed configurations are in Appendix ~\ref{app:archetypes}. Agents are initialised with one of three stance categories, Opposing, Neutral, or Supportive, using a balanced 1:1:1 distribution. The framework supports configurable population sizes through a scaling multiplier, allowing systematic expansion of the simulated society while preserving group composition. Agents interact over multi-dimensional policy topics extracted by an LLM-based Scenario Dispatcher. For reproducibility, all stochastic components—including personality initialisation, topology generation, and interaction sampling—are seeded deterministically. The system prompt is provided in Appendix~\ref{app:prompt}.




\subsubsection{Evaluation Metrics.}
Following prior work in opinion dynamics and multi-agent social simulation, we evaluate belief evolution using a set of trajectory-averaged metrics that capture polarisation, diversity, extremization, and behavioural dynamics. Polarization is measured as the variance of agent beliefs on the dominant topic $k^*$ at tick $t$, $\text{Pol}(t) = \mathrm{Var}(\{b_i^{k^*,t}\}_{i=1}^N)$, a standard variance-based polarization metric widely used in classical opinion dynamics models such as DeGroot~\citep{degroot1974reaching} and Friedkin–Johnsen~\citep{friedkin1990social}. 

To quantify opinion diversity, we compute Shannon entropy over five discretised belief bins, $H(t) = -\sum_c p_c \log_2 p_c$, where $p_c$ denotes the proportion of agents in bin $c$. Extremization is measured by the fraction of agents whose beliefs exceed a predefined extremity threshold, $\frac{1}{N}\sum_{i=1}^N \mathbf{1}(|b_i|\ge0.6)$, capturing the prevalence of radical opinions. We further compute the mean radicalisation level across agents and topics as $\bar{\mathcal{R}} = \frac{1}{N}\sum_{i=1}^N \frac{1}{K}\sum_{k=1}^K |b_i^k|$, which measures the overall ideological intensity of the population. In addition to belief-distribution metrics, we track behavioural and structural properties of the agent population. Personality diversity is measured as the mean variance of OCEAN personality traits across agents. 

We also compute rationale tag diversity, defined as the average number of distinct reasoning tags generated per agent, and reflection activity, measured as the mean number of reflection entries stored in each agent’s memory. Finally, for anchoring-enabled configurations, we report the mean anchoring strength $\rho$ and the mean anchor drift $\frac{1}{N}\sum_{i=1}^N |b_i - m_i|$, which measures the deviation between agents’ current beliefs and their intrinsic belief anchors.

\section{Macro-level Analysis and discussion}
This section investigates three fundamental research questions concerning collective opinion formation and topic evolution in simulated agent societies. 

\textbf{RQ1: Anchoring Dynamics.} 
To what extent do agents exhibit anchoring effects comparable to those observed in human decision-making? Specifically, we examine whether the anchoring mechanism embedded in our simulation produces belief adjustment patterns similar to empirically documented anchoring behaviour in human cognition.

\textbf{RQ2: Corpus-grounded effects on agent reasoning.}
To what extent does corpus-grounded initialisation influence how agents reason about controversial topics? Specifically, we investigate whether corpus-derived priors constrain agents to produce more fixed and homogeneous responses, or instead enable greater diversity in beliefs, framings, and interaction patterns.

\textbf{RQ3: How can social simulations quantify and regulate echo chambers and premature consensus?}
We study how bounded confidence and related interaction mechanisms shape polarization and diversity, and whether simulation outcomes can be used to infer real-world resilience to prior beliefs and effective bounded-confidence levels in controversial discourse.

\subsection{Emergent Phenomena}


\paragraph{Personality-dependent persuadability.}
Agents with high Openness ($O > 0.7$) and low anchoring strength ($\rho < 0.1$) change opinions 3--4$\times$ more frequently than agents with low Openness ($O < 0.3$) and strong anchoring ($\rho > 0.3$).
This heterogeneity in persuadability is essential for producing realistic group-level dynamics: opinion change occurs at the margins, driven by persuadable individuals, while the core of each group remains anchored. This phenomenon is really similar to the anchoring effect observed in humans' behavior~\cite{mcelroy2007susceptibility}. 



\subsection{Case Study}

\subsubsection{Case study 1: the overturning of the Roe v Wade decision that legalised abortion in the US in 1973.}

We instantiate \method{} on a policy debate scenario centred on the overturning of \emph{Roe v. Wade} and examine whether the resulting dynamics exhibit realistic post-shock opinion evolution. The full configuration produces persistent but bounded disagreement rather than collapsing to consensus: across runs, it maintains non-trivial polarization (variance: ($0.3559 \pm 0.003$); Esteban–Ray: ($0.1226$)), substantial stance diversity (($0.5958 \pm 0.002$)), and a bimodality coefficient of ($0.5605 \pm 0.001$), indicating a stable multi-cluster opinion structure rather than homogeneous convergence. At the same time, stance bias remains near zero (($-0.001 \pm 0.002$)), suggesting that the simulation does not drift toward an artificially one-sided aggregate outcome. high topical consistency (($0.980 \pm 0.006$)) ensures agents remain focused on the issue, while radicalisation (($0.493 \pm 0.001$)) indicates that the debate sustains committed subgroups without degenerating into universal extremity. 

\begin{figure*}[t]
    \centering
    \begin{subfigure}[t]{0.4\textwidth}
        \centering
        \includegraphics[height=0.2\textheight,width=\textwidth,keepaspectratio]{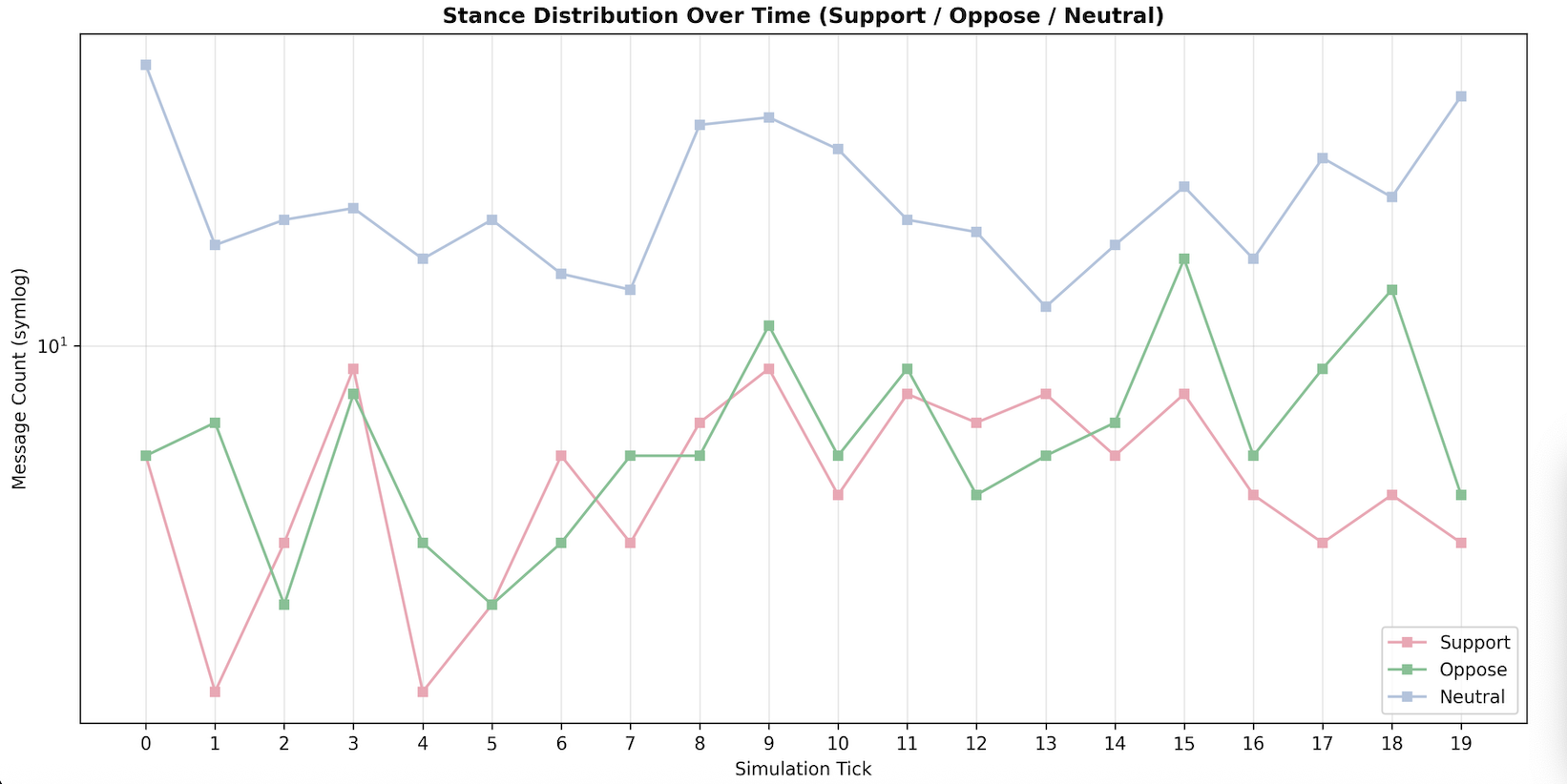}
        \caption{Stance distribution over time.}
        \label{fig:stance_distribution}
    \end{subfigure}
    \hfill
    \begin{subfigure}[t]{0.5\textwidth}
        \centering
        \includegraphics[height=0.2\textheight,width=\textwidth,keepaspectratio]{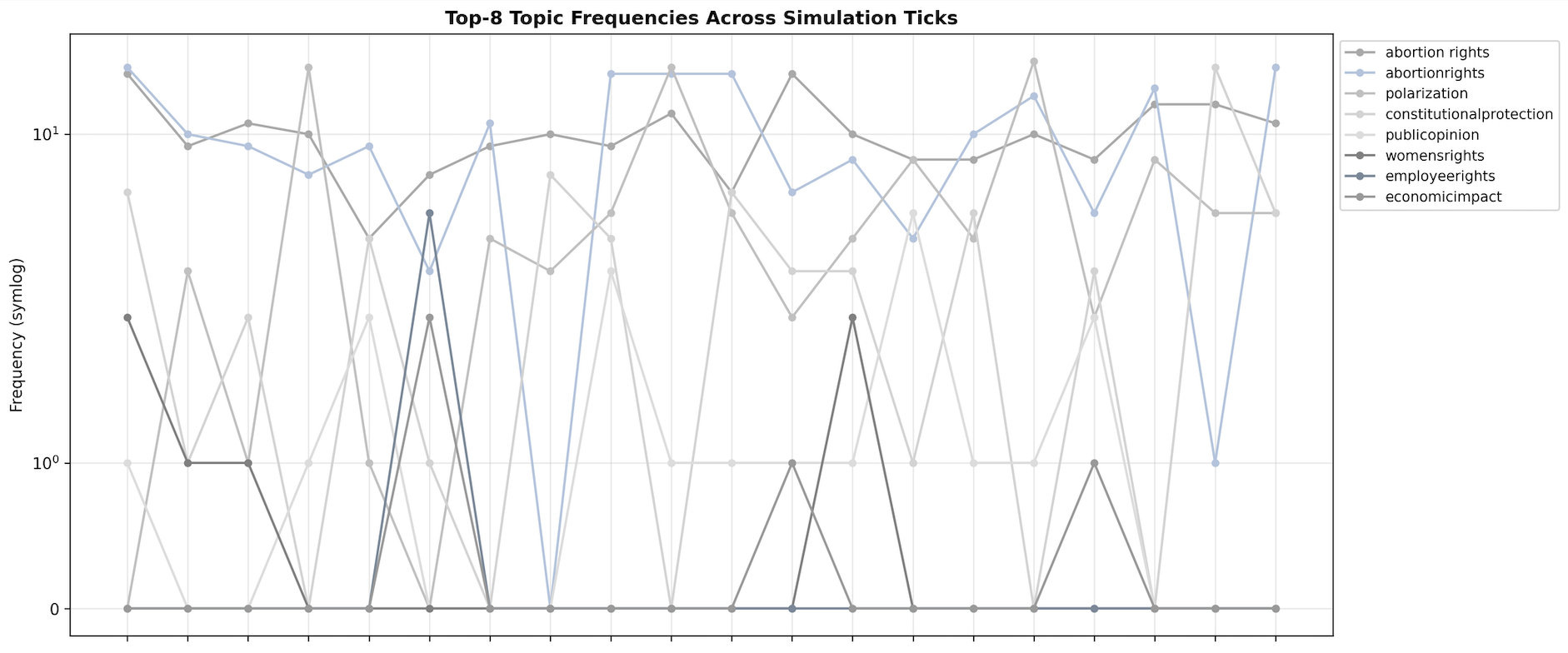}
        \caption{Top topic frequencies across simulation ticks.}
        \label{fig:emotion_distribution}
    \end{subfigure}

    \vspace{0.5em}

    \begin{subfigure}[t]{0.9\textwidth}
        \centering
        \includegraphics[height=0.26\textheight,width=\textwidth,keepaspectratio]{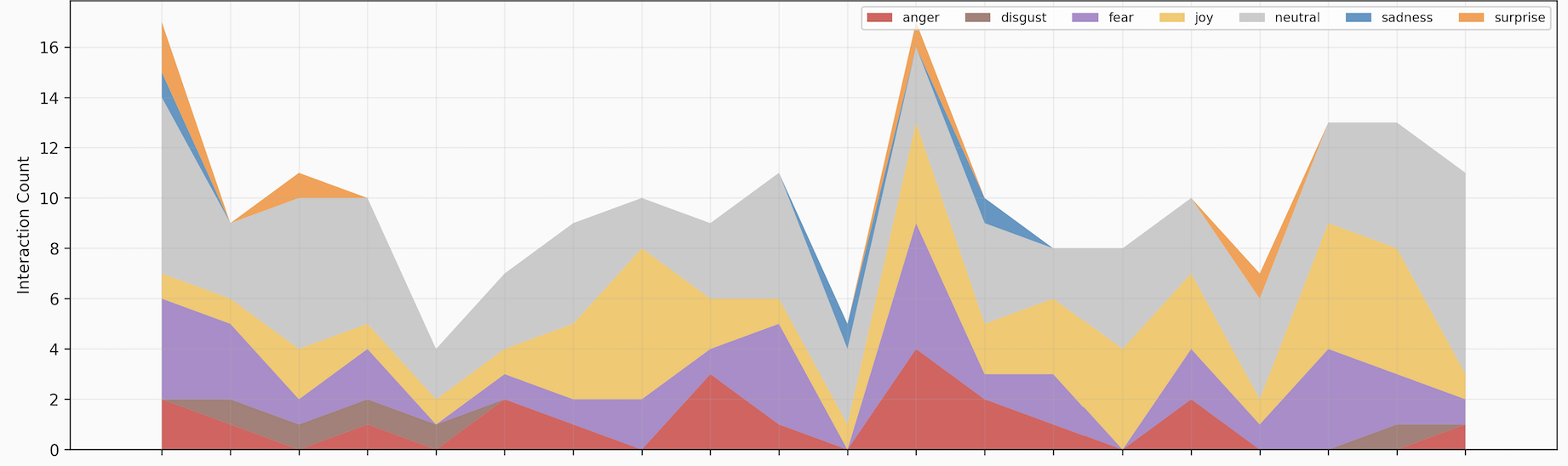}
        \caption{Top topic frequencies across simulation ticks.}
        \label{fig:topic_frequencies}
    \end{subfigure}

    \caption{Temporal dynamics of stance, emotion, and topic salience in the Roe v.~Wade case study.}
    \label{fig:roe_case_temporal_dynamics}
\end{figure*}

\begin{figure*}[t]
    \centering
    \includegraphics[width=0.7\linewidth]{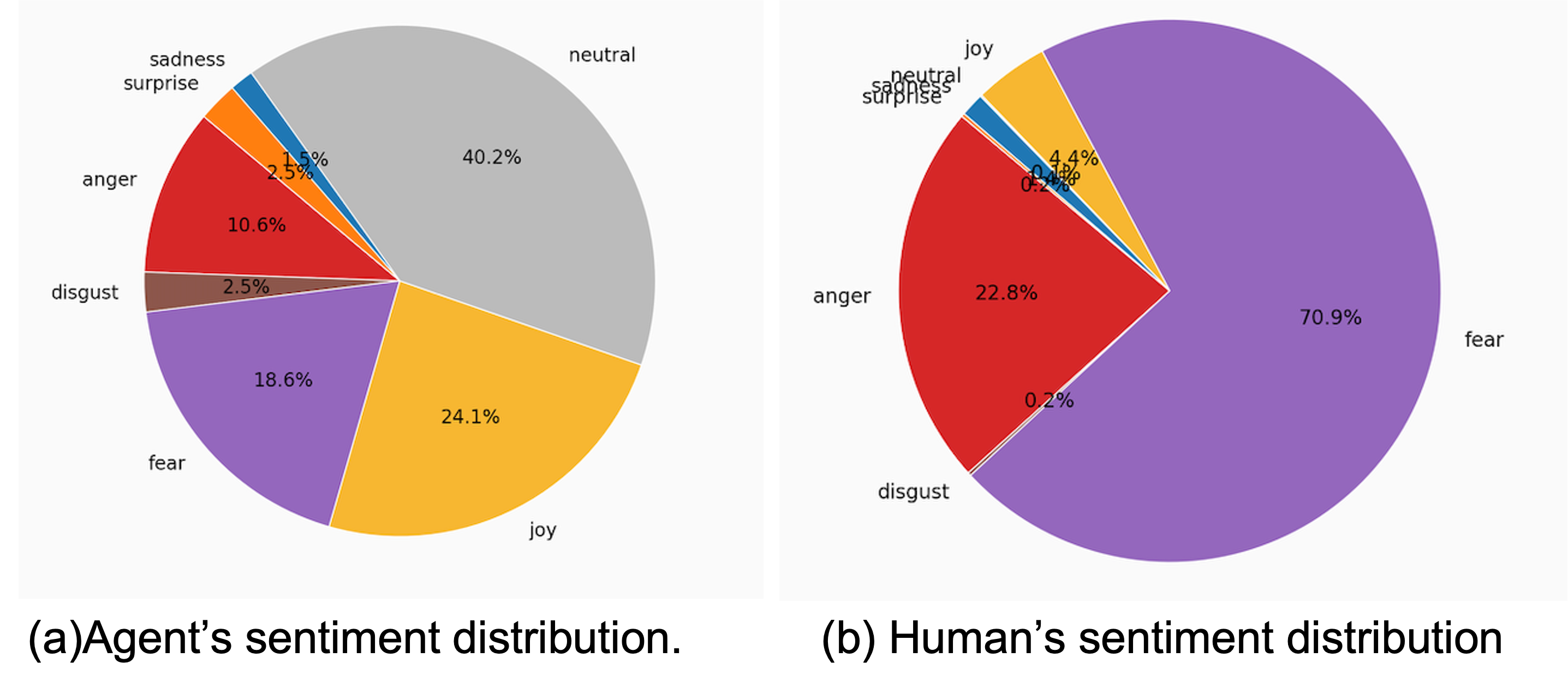}

    \caption{Comparison of the sentiment patterns of agents and humans in the simulation of the overturning of \textit{Roe v.~Wade}.}
    \label{fig:sentiment_comparison_result}
\end{figure*}

As shown in Figure~\ref{fig:roe_case_temporal_dynamics} and ~\ref{fig:sentiment_comparison_result}, support for the overturning of \textit{Roe v. Wade} drops sharply to a minimum around the first and fourth round of interaction, with overall agent sentiment shifting toward anger. We further compare the simulated outcomes with real-world social media data collected from discussions on the overturning of \textit{Roe v. Wade}\footnote{\url{https://www.kaggle.com/datasets/bwandowando/-roe-v-wade-twitter-dataset}}. The comparison shows that both the simulation and the empirical data exhibit highly polarised discourse, sustained disagreement, and strong affective responses, indicating that the simulation captures key qualitative patterns of real-world opinion dynamics.

However, a notable difference remains: while the agent-based simulation tends to produce relatively neutral and structured responses, real-world human discourse exhibits significantly stronger emotional intensity, with anger and fear dominating the majority of expressed sentiments. One of the reasons could be traced to the AI safety policy; once the agent responds to the tendency of extremization, it would freeze by the AI API supplier\footnote{\url{https://policies.google.com/terms/generative-ai/use-policy}}. Further details can be found in the topic deviation analysis, sentiment statistics, and original logs reported in Appendix~\ref{app:topic_emotion_mismatch} and Appendix~\ref{app:agent_logs}. 

Finally, we evaluate the effect of bounded confidence on polarisation in the \textit{Roe v. Wade} overturning scenario. Higher bounded-confidence thresholds ($>0.8$) lead to stronger echo chamber formation and significantly increased polarisation, whereas lower thresholds reduce polarisation and promote greater topic diversity across agents. By aligning these simulation outcomes with real-world controversial discourse, our simulation results provide an explainable model for inferring the population’s resilience to prior beliefs and effective bounded-confidence levels. 

\subsubsection{Case study 2: Australia social media ban}

This section reports an internal validity analysis of the simulation of Australia's social media ban policy influence. Consequently, the claims should be interpreted as \emph{simulation-internal evidence}: they describe the behaviour of the modelled agent society under the specified sampling and interaction assumptions, but do not by themselves constitute an empirical estimate of a real-world population effect.

\paragraph{Sampling Design and Analytical Population}
\label{subsec:2749c0df-sampling}


The simulation population was generated through stratified sampling over initial stance categories, while preserving role-group proportions where possible. The final full sample contained \(N=43\) agents, comprising 34 \texttt{CitizenGroup} agents, 3 \texttt{GovernmentGroup} agents, 3 \texttt{BusinessGroup} agents, and 3 \texttt{EducationGroup} agents. The realized stance distribution was mildly imbalanced, with 17 supportive, 14 opposing, and 12 neutral agents. We therefore treat the full \(N=43\) sample as the primary simulated population. In addition, we report results for a matched balanced subset of \(N=36\), containing 12 agents from each initial stance category, as a robustness check against stance-composition effects.

\begin{wrapfigure}{r}{0.35\textwidth}
    \centering
    \vspace{-0.8em}
    \includegraphics[width=0.35\textwidth]{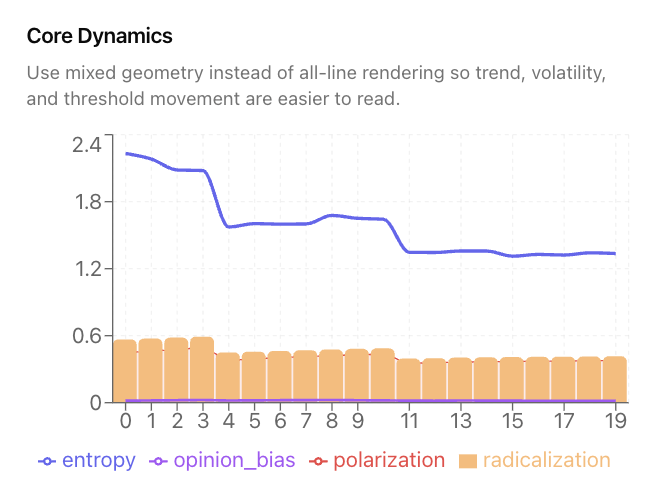}
    \caption{Core Opinion dynamics of simulation of Australia social media ban. }
    \label{fig:2749c0df-core-dynamics}
    \vspace{-1.0em}
\end{wrapfigure}

\begin{table}[t]
\centering
\caption{Sampling composition and stance-conditioned belief movement in the simulation of Australia's social media ban policy. Belief deltas are final mean belief minus initial mean belief within the corresponding initial-stance stratum.}
\label{tab:2749c0df-sampling}
\small
\begin{tabular}{llrrrr}
\toprule
\textbf{Population} & \textbf{Initial stance} & \textbf{Count} & \textbf{Initial belief} & \textbf{Final belief} & \(\boldsymbol{\Delta}\) \\
\midrule
Sampled population & Supportive & 17 & 0.1206 & 0.1706 & 0.0501 \\
Sampled population & Opposing   & 14 & -0.1665 & -0.1464 & 0.0202 \\
Sampled population & Neutral    & 12 & 0.0812 & 0.0395 & -0.0416 \\
\midrule
Balanced \(1{:}1{:}1\) cohort & Supportive & 12 & 0.1982 & 0.2796 & 0.0815 \\
Balanced \(1{:}1{:}1\) cohort & Opposing   & 12 & -0.1996 & -0.1919 & 0.0077 \\
Balanced \(1{:}1{:}1\) cohort & Neutral    & 12 & 0.0812 & 0.0395 & -0.0416 \\
\bottomrule
\end{tabular}
\end{table}


Figure~\ref{fig:2749c0df-core-dynamics} and Table~\ref{tab:2749c0df-sampling} together suggest that this simulation is better understood as a convergence-oriented belief-adjustment process rather than a polarization-amplifying one. At the macro level, entropy declines substantially, from approximately 2.3 at the beginning of the simulation to around 1.3 by the final ticks, indicating a reduction in aggregate opinion diversity. In contrast, polarization and opinion bias remain low and stable, while radicalization shows no sustained upward trend. The system therefore becomes more concentrated without clearly separating into increasingly polarized camps.

Table~\ref{tab:2749c0df-sampling} shows that this convergence is not produced by uniform movement across all stance groups. In the full sampled population, initially supportive agents move further in the supportive direction, increasing from 0.1206 to 0.1706. Initially opposing agents shift only slightly upward, from -0.1665 to -0.1464, suggesting limited moderation rather than further opposition. Neutral agents move modestly downward, from 0.0812 to 0.0395. The balanced \(1{:}1{:}1\) cohort shows the same pattern: supportive agents exhibit the strongest positive movement, opposing agents remain comparatively stable, and neutral agents shift slightly downward.

Overall, the results suggest that convergence occurs through different patterns of belief movement across initial-stance groups, rather than through uniform movement by all agents. The decline in entropy reflects reduced dispersion in the belief distribution, but the stable polarization trajectory indicates that this compression does not correspond to stronger ideological division. Since CitizenGroup agents dominate the sampled population, aggregate dynamics are primarily shaped by citizen-level interactions, while institutional groups should be interpreted as smaller but potentially influential actor classes. Finally, the balanced cohort improves descriptive comparability across initial stances, but it does not constitute randomized assignment; differences across strata should therefore be interpreted as modeled response heterogeneity rather than externally identified causal effects.

\paragraph{Causal Graph Evidence}
\label{subsec:2749c0df-causal-graph}

\begin{figure*}[t]
    \centering
    \includegraphics[width=0.95\linewidth]{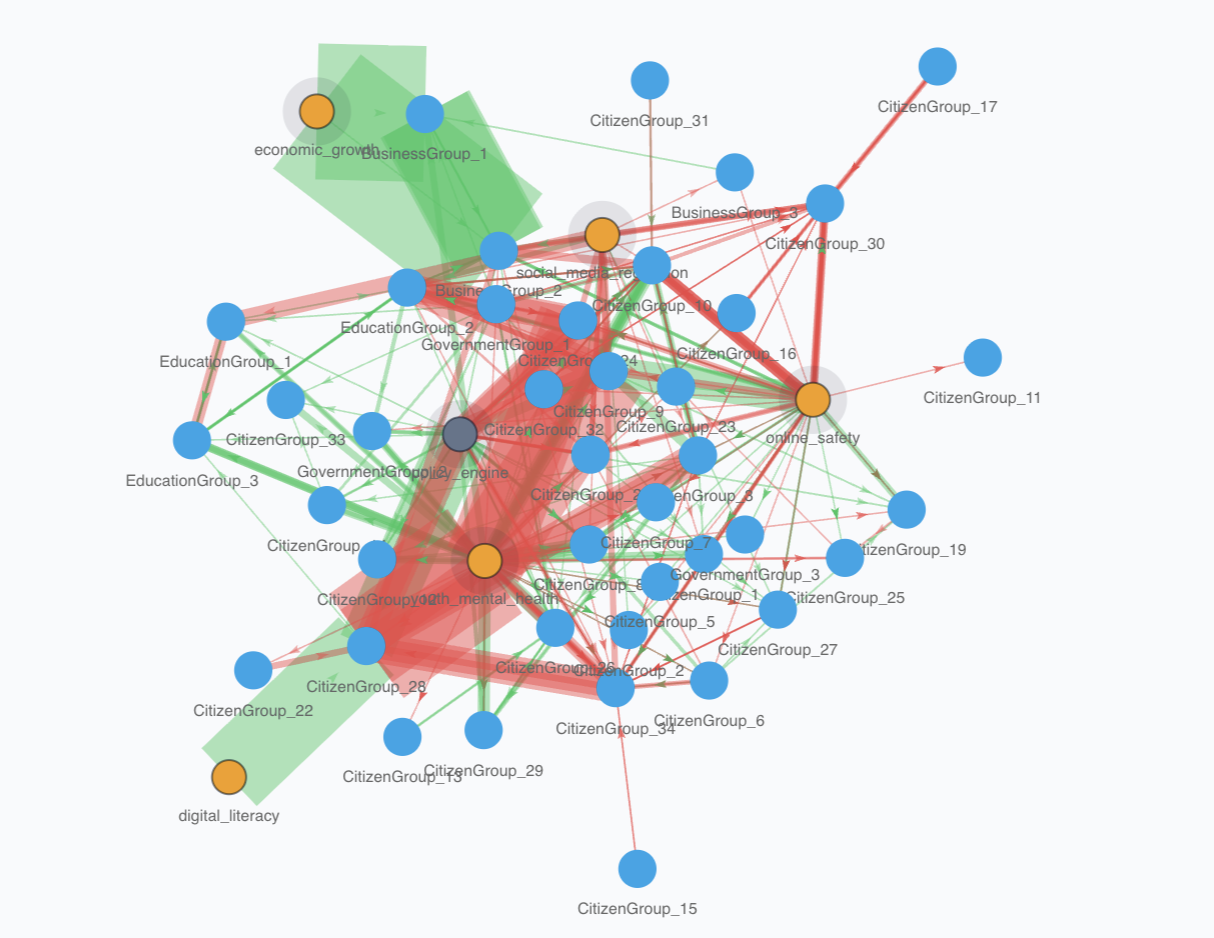}
    \caption{Agent-level causal graph for the Australia social media ban simulation. Nodes represent agents, topics, or source hubs involved in belief updates. Green edges indicate positive topic-to-topic influence, while red edges indicate negative topic-to-topic influence. }
    \label{fig:causal_effect_belief_shift}
\end{figure*}

Figure~\ref{fig:causal_effect_belief_shift} visualizes the agent-level causal graph derived from logged belief updates, message-shock updates, and topic-expansion events.  In this graph, blue nodes denote agents, orange nodes denote topic hubs, and the gray node denotes the policy-event source.  Green and red directed edges represent positive and negative belief movements respectively; edge thickness reflects the magnitude of simulated belief movement.  We use the term ``causal'' in the process-tracing sense: edges encode provenance relations in the simulation log, not externally identified causal effects in an observational population.

The graph contains a dense core around \texttt{policy\_engine}, \texttt{youth\_mental\_health}, \texttt{online\_safety}, and \texttt{social\_media\_regulation}.  This structure indicates agenda-setting concentration: although many agents participate in the simulated discourse, belief movement is organised around a small set of issue hubs.  In the underlying event log, the most frequent causal-event classes are agent inner belief updates (\(n=2851\)), communicative exposure effect updates (\(n=1953\)), and topic-expansion events (\(n=1785\)).  The derived agent causal graph contains 9557 event-level edges, while the group-level aggregation contains 4 group nodes and 69 group-level edges.

\begin{table}[t]
\centering
\caption{Dominant causal sources and issue targets by cumulative absolute belief movement in the simulation.}
\label{tab:2749c0df-causal}
\small
\begin{tabular}{llr}
\toprule
\textbf{Category} & \textbf{Node or topic} & \textbf{Cumulative \(|\Delta|\)} \\
\midrule
Source & \texttt{policy\_engine} & 3.1764 \\
Source & \texttt{EducationGroup\_2} & 1.2464 \\
Source & \texttt{CitizenGroup\_3} & 1.2108 \\
Source & \texttt{BusinessGroup\_2} & 1.0722 \\
\midrule
Affected agent & \texttt{CitizenGroup\_28} & 2.8932 \\
Affected agent & \texttt{BusinessGroup\_2} & 1.8061 \\
Affected agent & \texttt{BusinessGroup\_1} & 1.6529 \\
Affected agent & \texttt{CitizenGroup\_24} & 1.4897 \\
\midrule
Issue topic & \texttt{youth\_mental\_health} & 9.9708 \\
Issue topic & \texttt{online\_safety} & 4.2221 \\
Issue topic & \texttt{economic\_growth} & 3.0602 \\
Issue topic & \texttt{social\_media\_regulation} & 1.3626 \\
Issue topic & \texttt{digital\_literacy} & 0.9280 \\
\bottomrule
\end{tabular}
\end{table}

Three graph-level observations are especially important.  First, \texttt{policy\_engine} is the largest source of cumulative absolute belief movement, which suggests that top-down scenario events exert stronger influence than any single peer actor.  Second, \texttt{youth\_mental\_health} accounts for the largest topic-level movement, more than twice the movement associated with \texttt{online\_safety}.  The simulation therefore frames the debate less as a generic regulatory dispute and more as a youth-welfare controversy.  Third, the graph contains both positive and negative high-magnitude edges, visible as overlapping green and red bands around the central topic nodes.  This is consistent with the final stance distribution: the run does not collapse into one-sided persuasion, but instead channels agents toward opposing interpretations of the same core issues.

\paragraph{Interpretation Against Social-Science Criteria}
\label{subsec:2749c0df-criteria}

Against standard criteria in computational social science, the run yields a mixed but interpretable pattern.  Under a polarization criterion, the aggregate result is not evidence of escalating polarization because variance-based polarization declines.  Under a deliberative-diversity criterion, however, the result is unfavorable: entropy and stance diversity decline sharply, indicating narrowing of the opinion space.  Under an echo-chamber or local-convergence criterion, the evidence is strong: within-group polarization falls almost to zero, implying that agents become highly similar to others in their local role-conditioned environment.  Under a diffusion/adoption criterion, the policy does not become broadly adopted, as adoption rate falls from 0.4884 to 0.3209.  Finally, under an agenda-setting criterion, the causal graph strongly supports a centralized framing effect: \texttt{policy\_engine} and the \texttt{youth\_mental\_health}/\texttt{online\_safety} topic pair dominate belief movement.

The simulation suggests a more specific pattern: agents become more aligned within local interaction contexts, while neutral positions become less stable over time. At the aggregate level, however, the population neither converges to a single consensus nor separates into increasingly polarised camps. The process is therefore better understood as local convergence around salient issue frames, rather than as either consensus formation or unconstrained polarisation.

\subsubsection{Case study 3: the U.S. Presidential Election}

\paragraph{Top Belief-Change Edges and Voter-Level Support Reconstruction}

This subsection further examines simulation through two diagnostic views: the largest belief-change edges in the agent causal graph and the reconstructed voter-level support network. Together, these views help distinguish between issue-level persuasion dynamics and final candidate-support alignment.


The largest belief-change edges indicate that the strongest simulated opinion shifts were concentrated around economic insecurity and welfare-related policy frames. In particular, \texttt{inflation\_and\_cost\_of\_living} repeatedly appears as the dominant negative belief-shift driver across citizen, business, and education agents. The largest positive shift, by contrast, is associated with \texttt{healthcare\_affordability}, which produced a strong positive movement for \texttt{GovernmentGroup\_3} at tick 13. This suggests that economic pressure operated mainly as a destabilizing or oppositional frame, whereas healthcare affordability occasionally functioned as a coalition-building frame.


\begin{table}[htbp]
\centering
\small
\caption{Representative top belief-change edges in simulation \texttt{3833dc6c}.}
\label{tab:3833dc6c-top-belief-change-edges}
\setlength{\tabcolsep}{4pt}
\renewcommand{\arraystretch}{1.12}
\begin{tabularx}{\textwidth}{r p{2.7cm} p{2.7cm} X r X}
\toprule
\textbf{Tick} & \textbf{Source} & \textbf{Target} & \textbf{Topic} & \(\boldsymbol{\Delta}\) & \textbf{Evidence} \\
\midrule
13 & healthcare affordability & GovernmentGroup\_3 & Healthcare affordability & +2.0000 & belief change from agent's own response \\
7  & inflation / cost of living & BusinessGroup\_4 & Inflation and cost of living & -1.6431 & belief change from agent's own response \\
2  & inflation / cost of living & EducationGroup\_4 & Inflation and cost of living & -1.5556 & belief change from agent's own response \\
1  & inflation / cost of living & CitizenGroup\_9 & Inflation and cost of living & -1.5000 & belief change from agent's own response \\
13 & GovernmentGroup\_4 & GovernmentGroup\_3 & Healthcare affordability exposure & +1.4000 & communicative exposure effect \\
7  & EducationGroup\_3 & BusinessGroup\_4 & Inflation exposure & -1.1502 & communicative exposure effect \\
6  & inflation / cost of living & CitizenGroup\_9 & Inflation and cost of living & -1.1425 & belief change from agent's own response \\
14 & housing affordability & CitizenGroup\_19 & Housing affordability & -1.0000 & belief change from agent's own response \\
14 & inflation / cost of living & CitizenGroup\_19 & Inflation and cost of living & -1.0000 & subtopic inference \\
4  & inflation / cost of living & EducationGroup\_4 & Inflation and cost of living & -0.9896 & belief change from agent's own response \\
\bottomrule
\end{tabularx}
\end{table}

Two patterns are theoretically important. First, the top edges are not evenly distributed across all election topics; they cluster around household cost pressure, healthcare, housing, and institutional election procedures. Second, the strongest movers include both ordinary citizen agents and institutionally positioned agents, especially business and education actors. This implies that the simulation does not model persuasion as a purely mass-public process; instead, elite or semi-elite interpretive groups also absorb and retransmit issue shocks.

\paragraph{Voter-Level Support Reconstruction}

The voter-level reconstruction maps each citizen agent to a candidate preference and extracts the issue reasons supporting that preference. Among 32 citizen voters, 27 were reconstructed as supporting Democratic Party and 5 as supporting Republican Party. This corresponds to an 84.4\% Democratic support share and a 15.6\% Republican support share, with no undecided voters in the final reconstructed network.

\begin{table}[htbp]
\centering
\small
\caption{Final reconstructed party support among citizen agents.}
\begin{tabular}{lrr}
\toprule
Party & Voters & Share \\
\midrule
Democratic Party & 27 & 84.4\% \\
Republican Party & 5 & 15.6\% \\
\bottomrule
\end{tabular}
\end{table}

\begin{wrapfigure}{r}{0.35\textwidth}
    \centering
    \vspace{-0.8em}
    \includegraphics[width=0.35\textwidth]{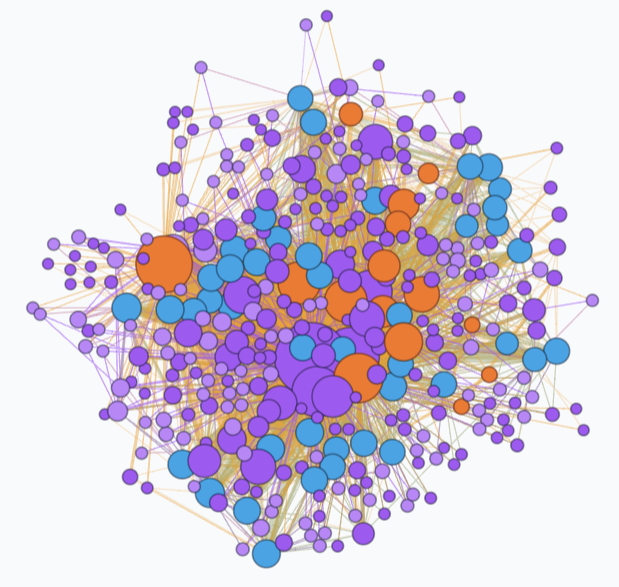}
    \caption{Topic Emergent network}
    \label{fig:topic_network}
    \vspace{-1.0em}
\end{wrapfigure}

The reconstructed support network shows Democratic Party Representative as the dominant attractor in the citizen-vote graph. However, this should not be interpreted as identical to the aggregate electoral-vote proxy. The individual-level voting network reflects reconstructed citizen preference, while the electoral proxy summarizes broader modeled viability under the case-study metric. Thus, the result should be read as strong simulated citizen-level alignment toward Democratic Party Representative, not necessarily as a direct prediction of electoral victory.

\begin{table}[htbp]
\centering
\small
\caption{Most frequent voter-level support reasons.}
\begin{tabular}{lr}
\toprule
Support reason & Count \\
\midrule
\texttt{inflation\_and\_cost\_of\_living} & 32 \\
\texttt{healthcare\_affordability} & 32 \\
\texttt{climate\_and\_energy\_policy} & 25 \\
\texttt{jobs\_and\_economic\_insecurity} & 15 \\
\texttt{vote\_for\_change} & 8 \\
\texttt{cost\_of\_living} & 5 \\
\texttt{climate\_action} & 4 \\
\texttt{affordable\_housing} & 2 \\
\texttt{healthcare\_for\_all} & 2 \\
\bottomrule
\end{tabular}
\end{table}

The support-reason distribution shows that voting was issue-driven rather than only party-label driven. All reconstructed voters cited inflation/cost of living and healthcare affordability, indicating that economic and welfare concerns formed the common evaluative baseline across the electorate. Climate and energy policy was also highly salient, appearing in 25 voter reconstructions, while jobs and economic insecurity appeared in 15. These patterns suggest that Democratic Party Representative's advantage was not simply produced by ideological sorting; it emerged from stronger alignment with the simulation's most active issue frames.

Several limitations should be noted when interpreting this 2024 U.S. election simulation. First, the simulation does not explicitly encode the full policy platforms proposed by each candidate during the campaign. It captures broad issue frames such as inflation, healthcare, climate, and employment, but not the complete set of candidate-specific policy proposals, campaign promises, or rhetorical shifts over time. Second, it does not model the geographically structured nature of the U.S. Electoral College. In particular, the simulation does not conduct separate state-level polling simulations or account for heterogeneous campaign effects across battleground and non-battleground states. 

The use of LLM agents as simulated voters also introduces representational limits. Large language models can approximate human subpopulations only when profile conditioning produces response distributions that resemble those of corresponding human groups~\citep{argyle2023out}. In this simulation, incomplete demographic, geographic, and partisan conditioning may leave the agents vulnerable to model-specific value priors, training-data biases, or assumptions skewed toward Western, Educated, Industrialised, Rich, and Democratic(WEIRD) populations. Such assumptions may overrepresent particular styles of political reasoning and underrepresent voters whose preferences are shaped by different institutional, cultural, regional, or socioeconomic contexts. The simulation should therefore be interpreted as an exploratory reconstruction of issue-based voter reasoning, not as a validated prediction of the 2024 U.S. Presidential Election.

Overall, the belief-change edge analysis and voter-level reconstruction converge on the same substantive interpretation: the simulated 2024 election discourse was organized primarily around material economic stress, with healthcare and climate operating as secondary but important coalition-building dimensions.  benefited from this issue ecology at the citizen-support level, while inflation-related shocks remained the principal source of belief volatility across the broader agent network.


\subsection{Ablation Results}
As shown in the picture ~\ref{fig:c1_c5_comparison} below, the anchoring component plays a critical role in sustaining realistic opinion dynamics. When anchoring is removed, agent updates become largely DeGroot-like, meaning that beliefs are driven primarily by neighbours’ opinions with little resistance to social averaging. This leads to overly rapid convergence, reduced persistence of individual belief states, and insufficient capacity to sustain meaningful topic evolution or stance differentiation over time. In particular, without anchoring, agents from different belief groups increasingly collapse toward a common neutral region, producing an unrealistic consensus-like outcome, which is also known as echo chamber mentioned by other works~\cite{zhang2025spark}. By contrast, adding anchoring in C5 yields the highest polarisation ($0.35 \pm 0.04$), restores a moderate level of extremization ($0.60 \pm 0.05$), and maintains stable radicalisation patterns ($0.56 \pm 0.04$), indicating that the model can preserve persistent yet bounded disagreement rather than collapsing into homogeneous neutrality. These results suggest that anchoring is essential for preventing over-smoothing in social belief updates and for producing belief persistence patterns that are more consistent with real-world controversial debates. 

\begin{figure*}[t]
    \centering
    \begin{subfigure}[t]{0.4\textwidth}
        \centering
        \includegraphics[width=\textwidth]{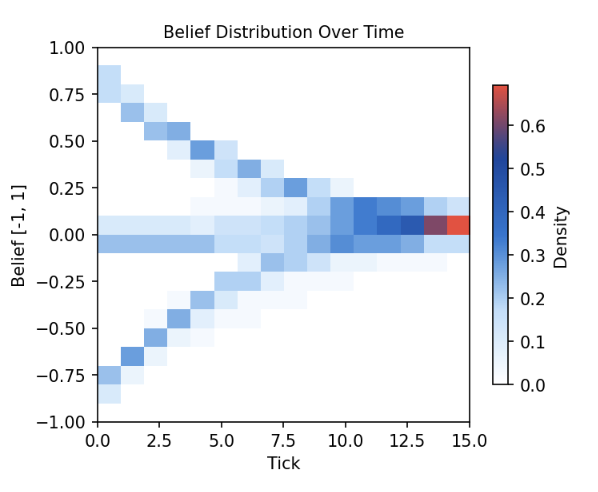}
        \caption{C1: Base}
        \label{fig:c1_result}
    \end{subfigure}
    \hfill
    \begin{subfigure}[t]{0.4\textwidth}
        \centering
        \includegraphics[width=\textwidth]{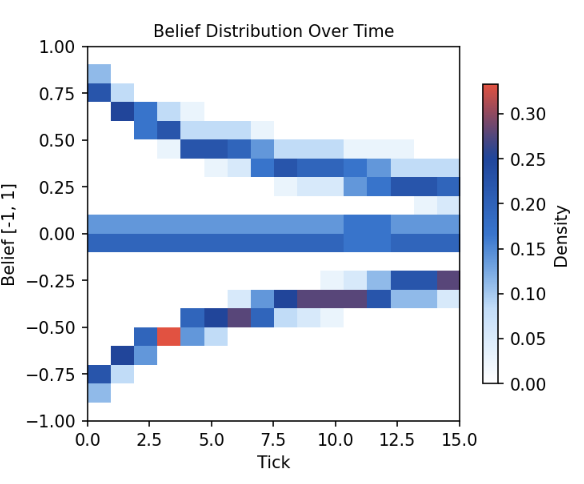}
        \caption{C5: +Anchoring}
        \label{fig:c5_result}
    \end{subfigure}
    \caption{
    Comparison of opinion dynamics under the base and anchored settings.
    C5 preserves more persistent subgroup disagreement and avoids the rapid neutral convergence observed in C1.
    }
    \label{fig:c1_c5_comparison}
\end{figure*}

\section{Limitations}
\label{sec:limitations}
Despite its strengths, this study still has several limitations. First, harmful discourse such as hate speech, toxic escalation, and targeted harassment is difficult to simulate faithfully with current LLMs. Safety alignment constraints suppress the generation of such content, while the contextual and emotional complexity of real-world hostile communication is hard to reproduce through prompting alone. Second, realistic opinion diversity requires far richer human-grounded data and likely bigger architectural changes than those implemented here. Although dynamic profiles improve over static personas, real human heterogeneity arises from lived experience, culture, class, identity, and long-term socialisation, not only from sampled personality traits and rationale templates. Capturing this more faithfully may require broader corpora of authentic human voices, longitudinal memory formation, and more structured cognitive representations~\cite{li2026simulating}.  In future work, we plan to explore the boundaries of these problems and provide solutions.



\section{Conclusion}
\label{sec:conclusion}

We presented \method{}, a cognitively grounded multi-agent social simulation framework that brings behavioral science principles into LLM-based social simulation. In \method{}, we propose a novel memory-anchored belief update mechanism, in which agents' accumulated experience serves as a cognitively meaningful anchor that shapes resistance to opinion change. This formulation enables \method{} to capture persistent belief trajectories consistent with the ``insufficient adjustment'' pattern reported in political psychology. Our layered memory architecture provides the experiential substrate for anchor formation, while dynamic corpus-grounded profiles equip agents with diverse identities, rationales, and internal states rather than static demographic labels.
Across five ablation studies, we demonstrates that each component contributes extensively to simulation realism, with full system producing emergent phenomena, including echo chambers, personality-dependent persuadability, memory-driven belief persistence, that qualitatively mirror patterns observed in real world. Overall, \method{} presents a cognitively grounded design offering a promising direction for building LLM-based social simulations that are both more stable and behaviorally realistic. Future work includes calibration against empirical opinion surveys, scaling to larger populations through batched inference, and extending the anchoring mechanism to model adaptive trust dynamics.

\section*{Reproducibility Statement}

We release \method{} as an interactive simulation system\footnote{\url{https://sciominds.vercel.app}} to support reproducible research in computational social science. 




\section*{Ethics Statement}
All corpus data are sourced from publicly available datasets, and do not disclose any personally identifiable information. All agents personas are synthetic, with demographic and psychological attributes designed solely for modelling social phenomena in a controlled and ethical setting. The code and simulation resources are intended strictly for research purposes, and all usage adheres to their designated scope.

\section*{LLM Usage Statement}
Large language models were used only for language editing and phrasing refinement during manuscript preparation. The authors verified all technical content, experimental details, results, and references, and take full responsibility for the submission.

\section{Acknowledge}
This research is funded by 
Centre for AI, Trust and Governance, The University of Sydney, Faculty of Arts \& Social Sciences, NSW 2006, Australia. 

\bibliography{paper/colm2026_conference}
\bibliographystyle{colm2026_conference}


\appendix
\section{Appendix}
\subsection{Group Archetype Configurations}
\label{app:archetypes}

Table~\ref{tab:archetypes_full} details the four group archetypes with OCEAN priors, communication styles, and trust structures.

\begin{table}[h]
\centering
\caption{Group archetype configurations. OCEAN values show group priors ($\mu \pm \sigma$).}
\label{tab:archetypes_full}
\small
\begin{tabular}{lcccccll}
\toprule
\textbf{Group} & $O$ & $C$ & $E$ & $A$ & $N$ & \textbf{Role} & \textbf{Trust Priorities} \\
\midrule
Citizen & $.60\pm.10$ & $.50\pm.12$ & $.50\pm.12$ & $.60\pm.10$ & $.50\pm.12$ & Public Forum & Education (0.6), Govt (0.4) \\
Government & $.50\pm.08$ & $.80\pm.08$ & $.40\pm.10$ & $.60\pm.08$ & $.30\pm.08$ & Policy Maker & Business (0.5), Edu (0.5) \\
Business & $.50\pm.10$ & $.70\pm.10$ & $.50\pm.10$ & $.50\pm.10$ & $.40\pm.10$ & Business & Government (0.6) \\
Education & $.70\pm.08$ & $.60\pm.10$ & $.50\pm.10$ & $.60\pm.08$ & $.40\pm.10$ & Research & Govt (0.5), Citizens (0.6) \\
\bottomrule
\end{tabular}
\end{table}

\subsection{Algorithm: Simulation flow}
\label{app:simulation_algorithm}

\begin{algorithm}[H]
\caption{Simulation Flow}
\label{alg:simulation_flow}
\begin{algorithmic}[1]

\REQUIRE Natural Language Scenario $S_{NL}$, Simulation ticks $T$, Agent distribution $\mathcal{A}$, Topic $K$, Corpus $\mathcal{C}$, Network Topology Config $\mathcal{G}$, Bounded Confidence Threshold $\theta_{BC}$, Anchoring Strength $\rho$
\ENSURE Belief trajectories $\mathbf{B}_{0:T}$, Polarization Metrics $\mathbf{M}_{0:T}$
\STATE $(K, T, \mathcal{G}_{groups}, \mathcal{E}_{policy}) \leftarrow \text{LLM\_ParseScenario}(S_{NL})$ 
\COMMENT{Extract: Topic $K$, Ticks $T$, Group Topologies $\mathcal{G}$, Policy Events $\mathcal{E}$}
\STATE $\mathcal{C}_K \leftarrow \text{LoadCorpus}(K, \mathcal{C})$ \COMMENT{Filter corpus for the specific topic}
\FOR{each agent $i \in \mathcal{A}$}
    \STATE $i.\text{profile} \leftarrow \text{InitializeProfile}(i.\text{group}, i.\text{stance}, \text{OCEAN\_traits})$
    \STATE $i.\text{memory} \leftarrow \text{SeedMemory}(\mathcal{C}, K, i.\text{profile})$ \COMMENT{Inject semantic priors}
    \STATE $b_i^{(0)} \leftarrow \text{ExtractInitialBelief}(i.\text{memory})$
\ENDFOR
\STATE $\mathcal{W} \leftarrow \text{BuildNetworkTopology}(\mathcal{A}, \mathcal{G})$ \COMMENT{e.g., Scale-Free, Small-World}
\STATE $\mathbf{W} \leftarrow \text{InitializeTrustMatrix}(\mathcal{W})$ \COMMENT{Row-normalized adjacency matrix}

\FOR{$t = 1$ \TO $T$}
    \STATE $\mathbf{b}^{(t-1)} \leftarrow \text{SnapshotBeliefs}(\mathcal{A})$ \COMMENT{Synchronous update state}
    \STATE $\mathcal{E}^{(t)} \leftarrow \text{DispatchPolicyEvents}(t)$ \COMMENT{Inject external news/shocks}

    \STATE \textbf{// 2.1: Information Propagation \& LLM Cognitive Processing}
    \FOR{each agent $i \in \mathcal{A}$}
        \STATE $\mathcal{M}_{i \leftarrow j} \leftarrow \text{ReceiveMessages}(\mathcal{W}, i, t)$ \COMMENT{Peer-to-peer sharing}
        \STATE $\mathcal{C}_{ctx} \leftarrow i.\text{memory}.\text{RetrieveContext}(\mathcal{E}^{(t)} \cup \mathcal{M}_{i \leftarrow j})$ \COMMENT{RAG via vector similarity}
        \STATE $p_i \leftarrow \text{AssemblePrompt}(i.\text{profile}, \mathcal{C}_{ctx}, i.\text{emotion}^{(t-1)})$
        \STATE $(m_i^{(t)}, \Delta b_i, e_i^{(t)}) \leftarrow \text{LLM\_Reasoning}(p_i)$ \COMMENT{Generate response, belief delta, emotion}
        \STATE $i.\text{memory}.\text{StoreWorkingMemory}(m_i^{(t)}, \Delta b_i)$
    \ENDFOR

    \FOR{each agent $i \in \mathcal{A}$}
        \STATE $S_i^{(t)} \leftarrow 0$, $\Omega_i \leftarrow 0$
        \FOR{each neighbor $j \in \mathcal{N}(i)$}
            \STATE $d_{ij} \leftarrow |b_i^{(t-1)} - b_j^{(t-1)}|$
            \IF{$d_{ij} > \theta_{BC}$}
                \STATE $\phi_{ij} \leftarrow 0$ \COMMENT{Sharp cutoff: Echo chamber wall (arXiv:2303.07563)}
            \ELSE
                \STATE $\phi_{ij} \leftarrow \max(0, 1 - d_{ij})$ \COMMENT{Homophily alignment}
            \ENDIF
            \STATE $S_i^{(t)} \leftarrow S_i^{(t)} + W_{ij} \cdot \phi_{ij} \cdot b_j^{(t-1)}$
            \STATE $\Omega_i \leftarrow \Omega_i + W_{ij} \cdot \phi_{ij}$
        \ENDFOR
        \STATE $S_i^{(t)} \leftarrow S_i^{(t)} / \max(\Omega_i, \epsilon)$ \COMMENT{Normalized social term}
        
        \STATE $A_i^{(t)} \leftarrow i.\text{memory}.\text{ComputeAnchor}()$ \COMMENT{Historical memory anchor}
        \STATE $b_i^{(t)} \leftarrow (1 - \rho) \cdot S_i^{(t)} + \rho \cdot A_i^{(t)}$ \COMMENT{Apply anchoring}
        \STATE $b_i^{(t)} \leftarrow \text{Clip}(b_i^{(t)}, -1, 1)$
    \ENDFOR

    \FOR{each agent $i \in \mathcal{A}$}
        \STATE $R_i^{(t)} \leftarrow \text{Reflect}(i.\text{memory}, b_i^{(t)} - b_i^{(t-1)})$
        \STATE $i.\text{memory}.\text{Consolidate}(R_i^{(t)})$ \COMMENT{Move to episodic/long-term memory}
    \ENDFOR
    
    \STATE $\mathbf{M}^{(t)} \leftarrow \text{ComputeMetrics}(\mathbf{b}^{(t)}, \mathcal{W})$ \COMMENT{e.g., Esteban-Ray Polarization}
\ENDFOR

\RETURN $\mathbf{B}_{0:T}, \mathbf{M}_{0:T}$
\end{algorithmic}
\end{algorithm}

\subsection{Prompt Template}
\label{app:prompt}

The system prompt for each agent follows this template:


\begin{quote}
\small
\begin{alltt}
You are a bounded agent in a social simulation. Respond as your character
with distinct opinions shaped by your unique background, rationale, and
rhetorical style.

=== AGENT PROFILE ===
- Role: \{role\}
- Group: \{group\}
- Personality: \{personality\_desc\}
- Style: \{style\_desc\}
- Goals: \{goals\}

=== CURRENT BELIEFS (Macro Stances) ===
\{beliefs\}
Each macro belief is an independent dimension. You may support one and
oppose another.

=== TOPIC PRIORS (Micro Subtopics) ===
\{subtopics\}
These represent the specific sub-issues you care about and how strongly
they influence your macro beliefs.

=== RELEVANT MEMORY ===
\{memories\}
(Refer to memories only if they are directly relevant to the event.)

=== INCOMING EVENT ===
\{event\}

=== EMOTIONAL STATE ===
\{emotion\}

=== CRITICAL RULES ===
1. DECOUPLED BELIEFS: Your beliefs are independent. Do NOT treat them as a
   single axis. You can support "abortion\_rights" while opposing "polarization".
2. TOPIC ACTIVATION: The event should activate specific micro subtopics, not
   just the macro belief. Update your macro belief based on which subtopics
   are triggered.
3. RATIONALE SPECIFICITY: Your core rationale is "\{rationale\_cluster\}".
   Ground your argument in this specific reason, not generic talking points.
4. RHETORICAL STYLE: Use "\{rhetorical\_style\}" framing throughout your
   response. Let your current emotional state drive your tone.
5. LANGUAGE DIVERSITY: Do NOT use generic phrases like "common ground",
   "stay informed", "everyday folks", "my family and community",
   "balanced approach", "evidence-based" (unless you are EducationGroup
   citing actual data). Write in a voice that is distinctly yours.

=== OUTPUT FORMAT (STRICT JSON) ===
You MUST output your response in valid JSON format exactly matching the
structure below. Do NOT output any text outside of the JSON object.

\{
  "message": "Your actual spoken/written response to the event. Keep under
  120 words. Be opinionated, authentic, and direct.",
  "active\_subtopics": [
    "List of subtopics from your priors that were triggered by this event"
  ],
  "emergent\_subtopics": [
    "Any new subtopics you are introducing to the discussion (optional)"
  ],
  "belief\_state": \{
    "topic\_key\_1": \{
      "stance": "Supportive|Neutral|Opposing",
      "score": <float between -1.0 and 1.0>
    \},
    "topic\_key\_2": \{
      "stance": "Supportive|Neutral|Opposing",
      "score": <float between -1.0 and 1.0>
    \}
  \},
  "belief\_composition": \{
    "topic\_key\_1": \{
      "subtopic\_1": <float weight 0.0-1.0>,
      "subtopic\_2": <float weight 0.0-1.0>
    \}
  \},
  "belief\_relations": [
    \{
      "source": "topic\_key\_1",
      "target": "topic\_key\_2",
      "relation": "positively\_linked|in\_tension|independent"
    \}
  ],
  "belief\_changes": [
    \{
      "type": "subtopic\_activation|belief\_reweighting|confidence\_change",
      "target": "topic\_or\_subtopic\_key",
      "old\_score": <float or null>,
      "new\_score": <float or null>
    \}
  ]
\}
\end{alltt}
\end{quote}

Group-specific user prompts further differentiate reasoning styles: Citizens focus on personal impact; Government agents emphasize strategic policy analysis; Business agents prioritize economic implications; Education agents ground arguments in evidence and research.



\subsection{Additional Analysis of Topic and Emotion Mismatch}
\label{app:topic_emotion_mismatch}

Although the simulation reproduces several high-level properties of polarised discourse, a notable gap remains between simulated and real-world expression. In particular, the agent-based outputs tend to be more neutral, structured, and deliberative, whereas real-world human discourse is substantially more affect-heavy and emotionally concentrated.

\paragraph{Sentiment mismatch.}
At the sentiment level, the simulation produces a comparatively balanced distribution, with negative, neutral, and positive labels accounting for $33.17\%$, $40.20\%$, and $26.63\%$, respectively. By contrast, the Twitter sample is overwhelmingly negative, with $97.0\%$ negative and only $3.0\%$ positive posts. This discrepancy is also reflected in the divergence metrics, where the weighted Jaccard similarity is $0.221$, the cosine similarity is $0.580$, and Jensen--Shannon divergence (JSD) reaches $0.525$. These results suggest that the simulation captures the existence of negative reactions, but substantially underestimates the extent to which real-world discourse collapses into strongly negative sentiment.

\paragraph{Emotion mismatch.}
A similar pattern appears at the emotion level. In the simulation, emotional expression remains relatively diffuse: fear accounts for $18.59\%$, anger for $10.55\%$, joy for $24.12\%$, and neutral affect for $40.20\%$. In contrast, the Twitter data are dominated by fear ($78.0\%$) and anger ($17.5\%$), with all other emotions appearing only marginally. The resulting weighted Jaccard similarity is $0.202$, cosine similarity is $0.411$, and JSD is $0.548$. This confirms that while the simulation can generate fear- and anger-related responses, it still produces overly moderated emotional distributions compared with real-world reactions to the \emph{Roe v. Wade} shock.

\paragraph{Topic overlap.}
Topic-level comparison further reveals limited lexical and topical overlap between simulated discourse and Twitter discourse. Even among the top-10 most frequent topics, only one topic (\textit{womensrights}) is shared, corresponding to a top-$k$ Jaccard score of $0.0526$. The cosine similarity of the full topic distributions is also low ($0.052$), while the JSD is high ($0.803$), indicating substantial divergence in topical salience.

\paragraph{Interpretation.}
Overall, these results suggest that the current simulation better captures structural properties of controversial debate---such as persistent disagreement and recurring issue focus---than the full emotional intensity and topic variability observed in real-world online discourse. In particular, the framework tends to produce more orderly and policy-oriented responses, whereas real-world human discourse in this case is far more emotionally concentrated, with fear and anger dominating the majority of expressions. This limitation should be taken into account when interpreting the realism of generated debate trajectories.

\begin{table}[t]
\centering
\caption{Comparison between simulation outputs and Twitter discourse for the \emph{Roe v. Wade} scenario.}
\label{tab:appendix_sim_twitter_compare}
\small
\begin{tabular}{lcc}
\toprule
\textbf{Metric} & \textbf{Simulation} & \textbf{Twitter} \\
\midrule
Negative sentiment & 0.3317 & 0.9700 \\
Neutral sentiment  & 0.4020 & 0.0000 \\
Positive sentiment & 0.2663 & 0.0300 \\
\midrule
Anger   & 0.1055 & 0.1750 \\
Fear    & 0.1859 & 0.7800 \\
Joy     & 0.2412 & 0.0250 \\
Neutral emotion & 0.4020 & 0.0000 \\
Sadness & 0.0151 & 0.0100 \\
Disgust & 0.0251 & 0.0050 \\
Surprise & 0.0251 & 0.0050 \\
\bottomrule
\end{tabular}
\end{table}

\begin{table}[t]
\centering
\caption{Distributional overlap between simulation and Twitter discourse.}
\label{tab:appendix_overlap_scores}
\small
\begin{tabular}{lccc}
\toprule
\textbf{Dimension} & \textbf{Weighted Jaccard} & \textbf{Cosine Similarity} & \textbf{JSD} \\
\midrule
Sentiment & 0.2207 & 0.5805 & 0.5246 \\
Emotion   & 0.2023 & 0.4111 & 0.5480 \\
Topic     & 0.0141 & 0.0521 & 0.8030 \\
\bottomrule
\end{tabular}
\end{table}

\subsection{Agent Logs}
\label{app:agent_logs}
\begin{table*}[htbp]
\centering
\caption{Examples of Emergent Affective Responses from Agents (Roe v. Wade Scenario)}
\label{tab:agent_emotions}
\renewcommand{\arraystretch}{1.3} 
\begin{tabularx}{\textwidth}{@{} l l l X @{}}
\toprule
\textbf{Agent Profile} & \textbf{Stance (Belief)} & \textbf{Emotion} & \textbf{Generated Narrative (Excerpt)} \\
\midrule

\textbf{Citizen\_3} & Pro-Choice ($+0.716$) & \textit{Anger / Outrage} & 
``I totally get the outrage from those pro-choice advocates—it's spot-on, and I'm right there with them on this one... the overturning of Roe v. Wade feels like a major blow to our constitutional protections... it could affect my family's access to healthcare decisions and make me worry about what other rights might slip away next.'' \\

\textbf{Citizen\_7} & Moderate ($+0.001$) & \textit{Fear / Anxiety} & 
``...that sounds pretty scary for women in my community, including friends who might worry about their options in tough situations... does it really change my family's access to healthcare or just stir up more fear? Let's not overreact; we need facts first.'' \\

\textbf{Citizen\_2} & Pro-Life ($-0.807$) & \textit{Frustration} & 
``Oh, man, this Supreme Court decision... is really throwing me for a loop—I'm not thrilled about it at all... right now, I'm frustrated that this could affect my family's future choices and even our cost of living if healthcare gets more complicated.'' \\

\textbf{Citizen\_5} & Pro-Life ($-0.743$) & \textit{Unease} & 
``...the Supreme Court's decision to overturn Roe v. Wade has me feeling pretty uneasy—but not for the reasons the pro-choice advocates are shouting about... It's got me pessimistic that we'll ever find common ground.'' \\

\textbf{Government\_1} & Pro-Choice ($+0.836$) & \textit{Deep Concern} & 
``This development is deeply concerning, as it undermines the principles of individual autonomy and equitable access to healthcare that our administration has consistently championed.'' \\

\bottomrule
\end{tabularx}
\vspace{1ex}
\small
\raggedright
\textit{Note:} Belief scores range from $-1.0$ (strongly Pro-Life) to $+1.0$ (strongly Pro-Choice). The narratives demonstrate how the agents' predefined profiles, initial memory seeds, and real-time social interactions trigger context-appropriate emotional arousal and vocabulary (e.g., ``outrage'', ``scary'', ``deeply concerning'').
\end{table*}

\end{document}